\documentclass[letterpaper, 10 pt, conference]{ieeeconf}  

\IEEEoverridecommandlockouts                              
\usepackage{graphicx} 

\title{\LARGE \bf
Does the D.C. Response of Memristors Allow Robotic Short-Term Memory and a Possible Route to Artificial Time Perception?*
}

\author{Ella Gale$^{1,2}$, Ben de Lacy Costello$^{1}$ and Andrew Adamatzky$^{1,2}$
\thanks{*This work was supported by EPSRC grant number EP/H01438/1}
\thanks{$^{1}$E. Gale, B. de Lacy Costello and A. Adamatzky are part of the Unconventional Computing Group, University of the West of England, Bristol, BS16 1QY, England
        { \{ella.gale,ben.delacycostello,andrew.adamatzky\}@uwe.ac.uk}}%
\thanks{$^{2}$E.Gale and A. Adamatzky are also affiliated with the Bristol Robotics Laboratory, Bristol, BS16 1QY, England
       }%
}

\begin{document}

\maketitle
\thispagestyle{empty}
\pagestyle{empty}

\begin{abstract}
Time perception is essential for task switching, and in the mammalian brain appears alongside other processes. Memristors are electronic components used as synapses and as models for neurons. The d.c. response of memristors can be considered as a type of short-term memory. Interactions of the memristor d.c. response within networks of memristors leads to the emergence of oscillatory dynamics and intermittent spike trains, which are similar to neural dynamics. Based on this data, the structure of a memristor network control for a robot as it undergoes task switching is discussed and it is suggested that these emergent network dynamics could improve the performance of role switching and learning in an artificial intelligence and perhaps create artificial time perception. 
\end{abstract}

\section{INTRODUCTION}

Robots need artificial perception to be able to plan for the future, learn from the past and make intelligent judgements in the present~\cite{Tani}, furthermore temporal concepts are needed for an agent to comprehend its environment and to successfully communicate with humans in a meaningful manner~\cite{Lingodroid}. Time perception covers several essential concepts which are needed for many tasks: duration of a task, perceived simultaneity of events with a small delta of time between them and ordering of events~\cite{Tani}. For example, to enable robots to switch tasks quickly between two (or more) different behaviours~\cite{Tani}, the robot needs some concept of different rules at different time states.

There has not been a great deal of work on artificial time perception and in studying human time perception we must turn to both neuroscience and philosophy. Robotics has drawn from these two areas, with a recent paper looking at the underlying structure of neural nets with regards to time-perception and rule switching plasticity~\cite{Tani}. Wittgenstein famously argued that a thought was impossible without a language containing that thought's concepts~\cite{Wittgenstein} and another recent paper looked at getting robots to develop their own language for time concepts~\cite{Lingodroid}, which interestingly involved errors due to individual robot's map not being entirely congruent with each other (a concept behind many human misunderstandings and conflict). In this paper, we will take the neuroscience view and consider the structure of an artificial `brain' that could understand temporal concepts.

In the brain, the time-perception tends to `ride along' with other mental processes: there is no part of the brain that is specifically evolved to deal with time perception~\cite{Tani} and different parts are associated with temporal aspects of behaviour on different timescales~\cite{TimePerception}. Instead, time seems to be perceived relative to internal neurological changes~\cite{Sumbre} as is evidenced by how it can be disrupted by disease~\cite{Alz}. The operation of time perception thus seems related to the network dynamics of the brain. 

The memristor is the 4$^{\mathrm{th}}$ fundamental circuit element~\cite{Chua1971} which is essentially a resistor with memory, and which, in 1971, was predicted (based on both electromagnetic and circuit theory) to be a two-terminal circuit element with a constitutive relation that would relate magnetic flux to charge. Although the constitutive relation technically covers everything about a circuit element's operation, this theory offered little clue of how to build such a device. Thus, the memristor was only related to an actual device in 2008~\cite{Strukov}, even though memristors had been previously experimentally studied and commercially investigated under the moniker of ReRAM. There are two different theories that model experimental memristor's operation: the phenomenological model~\cite{Strukov}, which is based on a 1-D model of variable resistors and which has been the basis of more complex models (such as those which include non-linear drift~\cite{94} or window functions) and many simulations (such as~\cite{84}); and the memory-conservation model (see ~\cite{F0} for a full description or~\cite{NM} for a summary) which is based on the electrodynamics of a 3-D model of variable resistors and fits with the constitutive relation. Recent simulations have shown that memristors can be used as synapses with artificial spiking neurons~\cite{David,STDP1}, theoretical results have demonstrated that action potential transport in real neurons can be modelled using memristors~\cite{Chua2013} and finally recent work~\cite{ICNAAM} has highlighted the memristors native spiking ability, all of which suggest memristors could be the basis of synthetic neuron analogues for use in an artificial brain.

In this short paper we will summarise some recent relevant memristor results and discuss how memristor networks might provide a route to incorporating time perception into an artificial brain/intelligence in a bio-mimetic or even human-like manner.

\section{MEMRISTOR'S SHORT-TERM MEMORY}

Memristors are commonly thought to be a.c. components. The pinched hysteresis loop used to identify a memristor is usually plotted in $V-I$ space, however this description is not complete without inclusion of the aspect of time-dependence. In a.c. systems this is evidenced as the dependence of the Lissajous curve lobe size on the voltage waveform frequency (and is why memristor papers now tend to include a graph showing this). This time dependence is due to the memory property of the device responding slower than the frequency of the voltage change (see~\cite{F0} for a discussion of what this memory property might be) and this memory property must have some characteristic time scale, $\tau$, associated with it that relates to the fundamental frequency, $\omega_0$ at which the a.c. voltage input produces the maximum hysteresis in the memristor $I-V$ curve. 

\begin{figure}[thpb]
      \centering
      \includegraphics[scale=0.4]{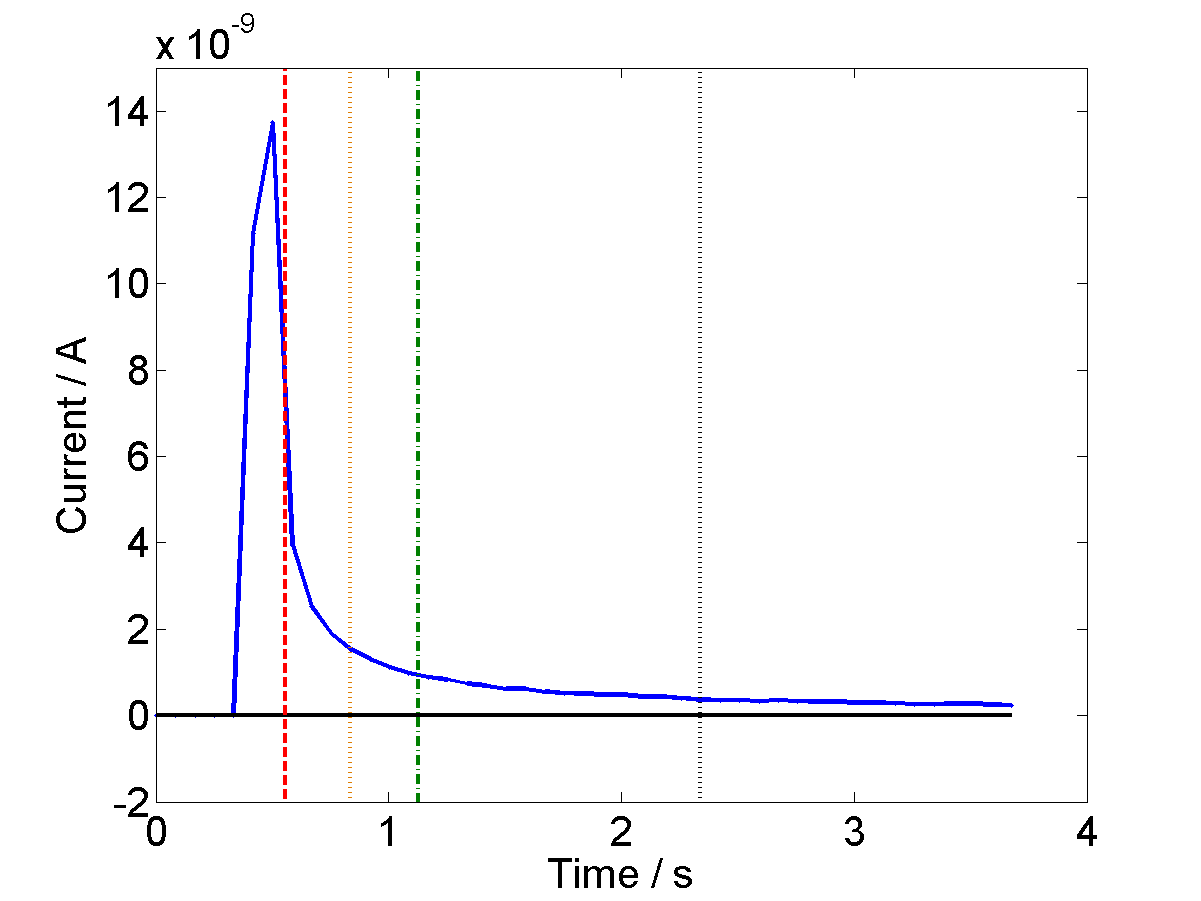}
      \caption{An example of a typical $I-t$ spike profile as taken from~\cite{ICNAAMJournal}. The characteristic timescale, $\tau$, is around 3.5-4s, marked on are the timescales for decay to a percentage of peak height: red line $\tau_{50}$ (decay to 50\%), orange line $\tau_{90}$, green $\tau_{95}$ and grey $\tau_{99}$.}
      \label{fig:ExampleSpike}
   \end{figure}

In steady voltage circuits (i.e. d.c. voltage input), we suggest that the time dependence is the commonly-observed current transients as seen when the voltage changes or is switched on or off, $\Delta V$. Figure~\ref{fig:ExampleSpike} taken from~\cite{ICNAAMJournal} shows an example current spike response to a step voltage. The characteristic timescale is related to the time taken for the spike to decay. As $\Delta V \rightarrow \delta V$ and we go from d.c. steps to an a.c. smooth curve with a set frequency, we can see that this characteristic timescale response is related to the hysteretic lag.

Whilst the memristor is responding to a voltage change, it is in a different state to a memristor that has responded: this is a form of memory. Specifically, it is a short term memory rather than a long-term memory or stateful response, this short-term memory could be used as a form of working memory. If a second spike is input into the system, it reacts differently as a result of the previous spike if and only if the second spike is happens within time $\tau$. 

The memory-conservation model of memristance~\cite{F0} introduces the concept of a second charge carrier, the ionic charge carrier, in addition to the electrons. This ionic charge carrier has a different mobility, speed and inertia to electrons and therefore takes longer to respond to voltage changes. Thus, in~\cite{F0} it is claimed that the lag which causes the memristor hysteresis under a.c. is due to the slower response time of the ionic charge carriers, and this is investigated in a forthcoming paper. The slow ionic charge response is apparent in the d.c. response as the decay of the $I-t$ curve. The characteristic timescale is a measure of the ionic charge carrier's slower response to a voltage change (which we expect will be related to it's ionic mobility) and it measures when the ionic effect is negligible. Therefore, if a second spike is received within time $\tau$, the ionic charge carrier hasn't recovered and responds at a different level than would be otherwise expected. 

\section{NETWORKS OF MEMRISTORS}

\subsection{SIMULATION}

What could such a short term memory be used for? A recent simulation based on the experimental observations outlined earlier showed that networks of memristors can learn, change and adapt~\cite{Mu0}. Memristor networks were designed and simulated with the purpose of composing and performing music. Each network node was a note (in terms of function) and a source drain or sink (in terms of modelled component), each connection a transition from one note to the other (function) and a pair of antiparallel memristors (component). The normalised conduction profile of a memristor under d.c. voltage as modelled using the memory-conservation model~\cite{F0} was descretized and used as a look-up table of how connection weight changed each time a connection was used.  

The network was capable of being seeded to produce music similar to the seed genre. Crucially, the memristor network continued to change and adapt as it was used. This shows a similar plasticity to the brain and a very simple type of Hebbian learning. This is different to evolutionary techniques because the learning is a result of using the networks rather than being directed by similarity to a fitness function or desired output. 

Although this experiment serves to demonstrate how a composing memristor machine could be built, there are similarities in the structure to the brain, which is also a learning network. It has been suggested that building a creative computer could require the almost accidental building of a brain-like computer.

\subsection{EXPERIMENT}

   \begin{figure}[thpb]
      \centering
      \includegraphics[scale=0.4]{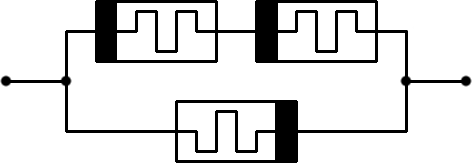}
      \caption{The three memristor circuit used to generate the output in figure~\ref{fig:Brainwave}}
      \label{fig:ThreeMems}
   \end{figure}

   \begin{figure}[thpb]
      \centering
      \includegraphics[scale=0.4]{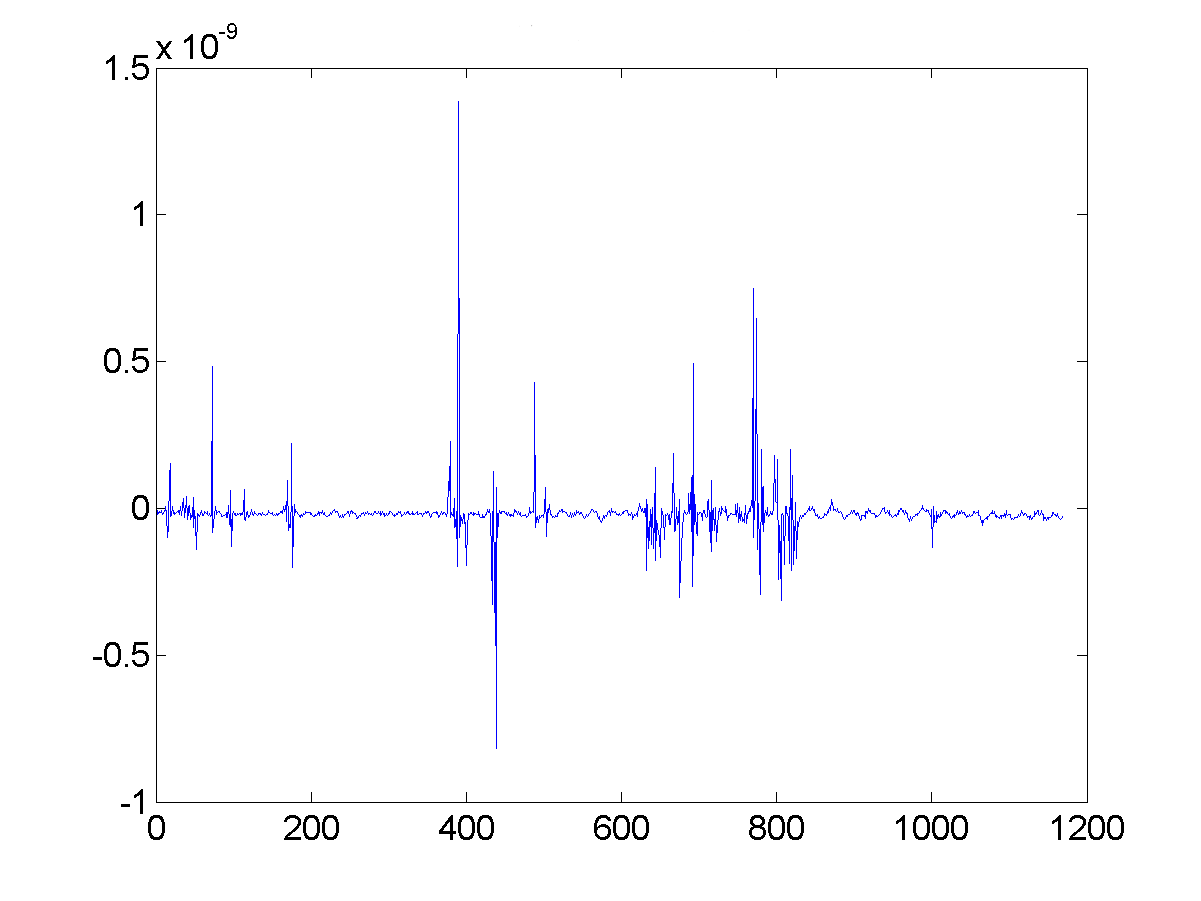}
      \caption{Brainwave like oscillations and spike trains that emerged from the circuit in figure~\ref{fig:ThreeMems}, as taken from~\cite{Mu0}. Utilising these dynamics could provide a route build a neuromorphic control computer for a robot.}      \label{fig:Brainwave}
   \end{figure}

Experimental data of highly simplified TiO$_2$ sol-gel memristor networks show some intriguing similarities to the dynamics of the brain. A simple `network' of just three memristors (arranged as in figure~\ref{fig:ThreeMems}) put under a constant positive d.c. voltage shows a current response similar to that shown in figure~\ref{fig:Brainwave}. Here we see sudden and large spiking responses against a background of an emergent oscillation, this is interesting as it even includes the measurement of negative currents under a positive driving voltage. These oscillations may arise due to the interaction within the network of the spiking components. As the brain also consists of spiking components and shows emergent mass synchronisation across certain frequencies (i.e. brainwaves) this result could show that we are on the right track in attempting to make neuromorphic (brain-like) computers with memristor networks. As complex behaviour (and learning) is seen in other networks of individually `simple' components, such as \textit{Physarum polycephalum}, a eukaryotic mould that can perform simple learning via interaction of its many nuclei, it suggests that the network structure may be of more importance than the precise components or measurables.  

Regardless of where these brain-wave-like dynamics emerge from, they have a use regarding time-perception. These oscillations can be used as an internal clocking signal, in fact, there is some evidence that this may be part of what synchronised spiking responses might be used for within the mammalian brain~\cite{Sumbre}. 

\section{DESIGN IMPLICATIONS FOR ROBOTIC CONTROL SYSTEMS}

Let us imagine a robot `brain' built from a complex memristor network in order to discuss how order and task switching might be encoded. Consider a standard test of asking a robot to navigate a T-maze under two different reward states, namely: I. reaching the left hand top of the `T' cross-bar; II. reaching the right-hand top of the `T' cross-bar (see~\cite{David,Tani} as relevant examples). We shall assume that after training different groups of memristors spike in different ways for the two solutions (as was seen with artificial neural networks in~\cite{Tani}). If the robot was operating under the rule turn left, we could see that continual brain-wave activity across the network could be used to keep the memristors in the correct short-term memory state for the robot to respond to external stimuli by turning left. Should we switch the reward state, we would expect the spike patterns to change (as seen in~\cite{Tani}), causing the memristor network to switch to the other behaviour. If pre-trained and plastic with distributed activity (i.e. the cause of the brainwaves) we can see that this function would allow the whole network to be switched due to the oscillations across it rather than waiting for each memristor to switch in turn as the robot processes the rule change and this may cause the robot to switch rules faster. This mechanism also allows the robot brain to have greater plasticity, as previous mechanisms can be co-opted to encode different responses applicable to situations outside those the robot has been trained for. Thus, an artificial brain based on memristor networks may offer a time-perception functionality due to the plasticity, learning and synchronisation properties of the network.




\section*{ACKNOWLEDGMENT}

E.G. thanks David Howard, Ioannis Georgialas and Oliver Matthews for helpful discussions.


\end{document}